# Fast Projective Image Rectification for Planar Objects with Manhattan Structure

J. Shemiakina[1,4], I. Konovalenko[3,4], D. Tropin[1,2,4], I. Faradjev[1,4]

[1] Federal Research Center "Computer Science and Control" of Russian Academy of Sciences, Moscow, Russia
[2] Moscow Institute of Physics and Technology (State University), Moscow, Russia
[3] The Institute for Information Transmission Problems of Russian Academy of Sciences, Moscow, Russia
[4] Smart Engines Service LLC, Moscow, Russia

## ABSTRACT

This paper presents a method for metric rectification of planar objects that preserves angles and length ratios. An inner structure of an object is assumed to follow the laws of Manhattan World i.e. the majority of line segments are aligned with two orthogonal directions of the object. For that purpose we introduce the method that estimates the position of two vanishing points corresponding to the main object directions. It is based on an original optimization function of segments that estimates a vanishing point position. For calculation of the rectification homography with two vanishing points we propose a new method based on estimation of the camera rotation so that the camera axis is perpendicular to the object plane.

The proposed method can be applied for rectification of various objects such as documents or building facades. Also since the camera rotation is estimated the method can be employed for estimation of object orientation (for example, during a surgery with radiograph of osteosynthesis implants). The method was evaluated on the MIDV-500 dataset containing projectively distorted images of documents with complex background. According to the experimental results an accuracy of the proposed method is better or equal to the-state-of-the-art if the background occupies no more than half of the image. Runtime of the method is around 3ms on core i7 3610qm CPU.

## 1. INTRODUCTION

Increase of computational power allows solving many difficult computer vision problems on mobile devices. Amongst others there is a task of projectively distorted image analysis. Most commonly this task includes object localization and rectification step. It is important for rectification to be metric so the angles and length ratios of the original object are preserved. Otherwise a refined object image would be corrupted with some nonprojective distortions that may result errors after further analysis, for example text recognition.

If an object template is known and contains many static elements the object can be located and rectified simultaneously with local features [1, 2]. In case of a rectangular object with a poor or unknown template we can detect its borders [3, 4, 5] and then calculate homogrpahy from the found quadrangle to a rectangular with known aspect ratio. However, this approach tends to misclassify inner segments as borders so rectified image may be stretched.

The localization problem in undistorted images is well-known [6, 7] so we can isolate the problem of rectification. We assume that object's inner structure belongs to the Manhattan World i.e. the majority of line segments has two dominant orthogonal directions. In this case common approach contains two steps: estimation of two vanishing points corresponding to the object directions and calculation of a rectification homography. A vanishing point might be found as an intersection of lines in an image. But due to data noise we have to construct a function of lines which has extrema in a vanishing point. Papers use many different approaches to select lines and to construct the function.

For lines to be detected Hough Transform [8] is widely used [9, 10, 11, 12]. In [9] vanishing points are estimated robustly as sine waves in Hough space. Paper [10] presents a method which chooses the point with the highest number of peaks of a skew projective histogram. Authors of [11, 12] present a neural network architecture based on a new FHT layer. It should be noted that in many cases when HT lines are employed the optimization function is calculated not on the image plane. But the image plane is usually preferred because the noise originates in the image [13].

Papers [14, 15] consider documents rectification based on text lines and vertical strokes of characters. They are extended to be lines and an optimization function is defined as a sum of squared distances from the lines to the sought vanishing point. Here one calculates consistency of a line and a vanishing point on the image plane. However, this approach ignores the fact that the farther the lines intersection is from the image center the higher noise it may have due to line noise. So the consistency value depends on that distance.

This problem can be avoided if one uses a model of data noise while constructing the optimization function. As there is no convenient model of line noise we can employ segments with Gaussian noise of their endpoints. This idea is used in [16, 17] where the consistency of a segment and a vanishing point is calculated as a squared distance from its endpoint to a line through the vanishing point and the segment middle point. But there is no proof of optimality of such an optimization function.

There are several different methods for rectification homography calculation with two vanishing points. In papers [9, 12] authors find four intersection points of lines through each vanishing point and 2 opposite image corners and transform them to the image corners. In [10, 15] separate homographies are constructed for each vanishing point to be transformed to infinity. Both methods don't preserve length ratios. In papers [18, 14] the camera model is used to rectify an image. In [14] the method for metric rectification is presented but a rectified image may be flipped. Authors of [18] estimate rectification based on the camera rotation for object to be perpendicular to the camera axis. However, they assume that the camera focal length equals 1 so the method also does not preserve length ratios.

In this work we propose an algorithm for vanishing point detection based on an optimization function proved to be optimal with the segment noise model described above. For image metric rectification we introduce an original algorithm that uses estimation of camera rotation and avoids the refined image flip. Also we discuss how to perform it with the unknown camera focal length.

The proposed method can be used for rectification of document or facade images without any information about objects locations or their templates. Also it can be applied while increase the spatial resolution in computer tomography. For that purpose in [19] authors place asymmetrically reflecting crystals behind the object. But the presence of these optical elements in the optical path leads to projective distortion of the recorded projections, which must be adjusted before sending the measured projections for reconstruction. If the object under investigation has Manhattan structure the proposed method is an adequate tool for the necessary correction.

Moreover, while we estimate camera rotation it can be interpreted as the object orientation. It can be useful in many computer vision tasks of object analysis, for example camera calibration or estimation of an osteosynthesis implant pose with a radiograph during a surgery.

## 2. CAMERA MODEL

The projective distortions in images can be described using the pinhole camera model (Fig. 1). Let $C$ be the camera center and $CXYZ$ be the camera coordinate system, where the axis $CZ$ is the camera optical axis. Then the image plane can be expressed as $Z = f$, and $f$ is the focal length. On the image plane we define a homogeneous coordinate system $Oxy$ so that the axes $Ox$ and $Oy$ are codirectional to $CX$ and $CY$ respectively. We assume that the principal point $p$ in which the optical axis and the image plane intersect is known and pixels are square. Then we can define the camera calibration matrix $K$ [13] that transforms points in the camera system to points on the image plane $Oxy$:

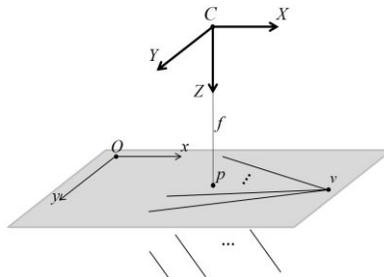

Figure 1. The camera and image coordinate systems, parallel line in 3D world, their projections and the corresponding vanishing point

$$K = \begin{pmatrix} f & 0 & p_x \\ 0 & f & p_y \\ 0 & 0 & 1 \end{pmatrix}$$

Let us consider $n$ parallel lines in the camera system. Their projections on the image plane will have a common point $v$ called a vanishing point. It can be infinite only if the lines are parallel to the image plane. A Manhattan object contains segments of two main directions hence two vanishing points should exist on its image.

## 3. VANISHING POINTS DETECTION

To find the vanishing point in the image we may detect the corresponding line segments. But all detectors can provide only noisy data. Thus we cannot simply intersect the lines passing through the segments. Then the sought point should be obtained by optimizing a function of the noisy segments. In this section we derive this function with Maximum likelihood estimation method.

### 3.1 Probabilistic model of a noisy segments set converging to a vanishing point

For the optimization function to be constructed we need to define the model of a noisy segments set on the image plane. Let us denote the sought vanishing point by $\mathbf{v} \in \mathbb{R}^2$. Let us consider the set of $n$ ideal line segments with endpoints referred to as $\mathbf{a}_i^1, \mathbf{a}_i^2 \in \mathbb{R}^2$, $i = 1..n$. If we denote the unit directional vector of $i$-th segment by $\mathbf{e}_i \in \mathbb{R}^2, \|\mathbf{e}_i\|_2 = 1$ the end-points can be described as $\mathbf{a}_i^j = \mathbf{e}_i t_i^j + \mathbf{v}$, $t_i^j \in \mathbb{R}$, $j = 1, 2$.

Then the end-points of noisy segments $s_i$, $i = 1..n$ can be considered as random variables $\mathbf{X}_i^j$ which are independently normally distributed: $\mathbf{X}_i^j \sim \mathcal{N}\left(\mathbf{a}_i^j, \Sigma\right)$ with the covariance matrix $\Sigma = \sigma^2 I_2$. Since the vanishing point and the ideal segments are unknown the distribution should depend on them. We denote all the parameters as $\theta \stackrel{\text{def}}{=} \{\mathbf{v}, \{\mathbf{e}_i\}, \{t_i^j\}, \sigma\}$. Then the probability density function of $\mathbf{X}_i^j$ can be derived: $f_{\mathbf{X}_i^j}(\mathbf{x} \mid \theta) = \frac{1}{2\pi\sigma^2} e^{-\frac{1}{2\sigma^2}\|\mathbf{x} - (\mathbf{e}_i t_i^j + \mathbf{v})\|_2^2}$

### 3.2 Optimization problem for a vanishing point estimation

Given a realization $\mathbf{x}_i^j$ of each random value $\mathbf{X}_i^j$ the log-likelihood function can be written as follows:

$$l(\theta) = -\log\left(\prod_{i=1}^n \prod_{j=1}^2 f_{\mathbf{X}_i^j}\left(\mathbf{x}_i^j \mid \theta\right)\right) = 2n\log\left(2\pi\sigma^2\right) + \frac{1}{2\sigma^2} \sum_{i=1}^n \sum_{j=1}^2 \|\mathbf{x}_i^j - (\mathbf{e}_i t_i^j + \mathbf{v})\|_2^2$$

The optimization problem is $\hat{\theta} = \arg\min_{\theta \in \Theta} l(\theta)$. Then sought vanishing point estimation can be obtained as:

$$\hat{\mathbf{v}} = \arg\min_{\mathbf{v}} \min_{\{\mathbf{e}_i\}} \min_{\{t_i^j\}} \min_{\sigma} l\left(\mathbf{v}, \{\mathbf{e}_i\}, \{t_i^j\}, \sigma\right) \quad (1)$$

We reduce the function $l\left(\mathbf{v}, \{\mathbf{e}_i\}, \{t_i^j\}, \sigma\right)$ by calculating an optimal value of parameter $\sigma$ while fixing another parameters and substituting it in the function. Similar procedure is performed for the parameters $\{t_i^j\}$ and the function $l\left(\mathbf{v}, \{\mathbf{e}_i\}, \{t_i^j\}\right)$. Then the vanishing point estimation problem can be expressed as:

$$\hat{\mathbf{v}} = \arg\min_{\mathbf{v}} \min_{\{\mathbf{e}_i\}} \sum_{i=1}^n \sum_{j=1}^2 \|(\mathbf{x}_i^j - \mathbf{v}) - \mathbf{e}_i (\mathbf{x}_i^j - \mathbf{v})^T \mathbf{e}_i\|_2^2 = \arg\min_{\mathbf{v}} \sum_{i=1}^n \min_{\mathbf{l}_i} \left(\rho^2(\mathbf{x}_i^1, \mathbf{l}_i) + \rho^2(\mathbf{x}_i^2, \mathbf{l}_i)\right) \quad (2)$$

where $l_i, i = 1..n$ are lines $\mathbf{x} = \mathbf{e}_i t + \mathbf{v}, t \in \mathbb{R}$ and $\rho(\mathbf{x}, \mathbf{l})$ is the distance between a point $\mathbf{x}$ and a line $\mathbf{l}$. That means that the consistency of a vanishing point $\mathbf{v}$ and a noisy segment $s_i$ equals to the distance between $s_i$ and an optimal line $\mathbf{l}_i$ through the $\mathbf{v}$ point. To estimate the directional vectors $\mathbf{e}_i$ of the optimal lines $\mathbf{l}_i$ we can use the Principal component analysis. Then the resulting optimization problem for a vanishing point estiamation can be expressed as:

$$\hat{\mathbf{v}} = \arg\min_{\mathbf{v}} \sum_{i=1}^{n} \lambda_{\min}(K_i(\mathbf{v})) \tag{3}$$

The problem cannot be solved explicitly so in the next Section we propose a numerical method for vanishing point estimation based on the derived optimization function.

### 3.3 Vanishing points detection on the image plane

We estimate the position of a vanishing point for the noisy set of line segment numerically. Thus the consistency measure $D(\tilde{v}, s)$ of a point-candidate $\tilde{v}$ and the segment $s$ should be formulated. According to (3) the consistency measure equals minimum eigenvalue of the matrix $K$: $D(\tilde{v}, s) = \lambda_{\min}(K_i(\mathbf{v}))$. However, the used model of a segments set does not imply the presence of outliers. Then let us say that if the consistency measure exceeds a predefined threshold $T_D$ then the segment is assumed to be an outlier for the point $\tilde{v}$. The new optimization problem is robust:

$$\hat{\mathbf{v}} = \arg\min_{\mathbf{v}} \sum_{i=1}^{n} \min(T_D, \lambda_{\min}(K_i(\mathbf{v}))) \tag{4}$$

The resulting output of the algorithm should be a pair of orthogonal vanishing point. Thus in this Section we construct the set of probable candidate points from which a pair could be chosen. First of all we obtain the set $E$ of rough candidates. It contains intersections of the segments which are long enough. The length threshold $T_L$ depends on the mean length of the segments. For each rough candidate $c$ we obtain the inliers set. Then the candidate set is inspected for repeated points. If two points have similar inliers the point with fewer of them is discarded from $E$.

Each rough candidate $e \in E$ is used as an initial point to evaluate numerically an optimal solution $e^*$ of problem (4). The optimization is conducted using the conjugate gradient method for the inlier segments of $e$. For each accurate candidate $e^*$ we compute the inliers set. Then we inspect them for repeated points and also discard candidate vanishing points which are located too close to the principal point $p$ (in which the optical camera axis and the image plane intersect). It allows eliminating wrong candidates corresponding to too strong projectivity. After this discarding procedure we obtain the resulting set $E^*$ of candidate vanishing points.

### 3.4 Selection of two orthogonal vanishing points

First of all we should derive the requirements for two candidate vanishing points to be orthogonal. If the positions of two vanishing points $v_1$ and $v_2$ are given absolutely accurate we can check their orthogonality explicitly. In case of the known camera focal length $f$ the calibration matrix $K$ defined in Section 2 is also determined. Let $v$ be a vanishing point on the image plane defined in the homogeneous coordinate system $Oxy$. A vector $V = K^{-1}v$ defined in the camera system is a directional vector of the parallel lines corresponding to the vanishing point. Thus for vanishing points $v_1$ and $v_2$ to be orthogonal the angle between corresponding vectors $V_1$ and $V_2$ also must be right.

But often the focal length $f$ is undefined. If the vanishing points are orthogonal we can calculate $f$ from the above: $f = \sqrt{-v_{1x}v_{2x} - v_{1y}v_{2y}}$. This value is computable only if the angle α between vectors $\overrightarrow{pv_1}$ and $\overrightarrow{pv_2}$ is greater than 90º. However, the vanishing points are estimated approximately. Due to inaccuracy the angle α on nondistorted images may be slightly less than the right angle. So the resulting constrain may be formulated as $\alpha > 90º - T_\alpha^1$.

Wider angles α between vanishing points lead to higher level of projective distortion. Images with very strong projectivity are hardly readable and thus not used for object analysis so we can constrain the vanishing point pairs to have angles less than a threshold $T_\alpha^2$.

Having the restrictions on vanishing point pairs we should choose one pair from the set of vanishing point candidates $E^*$. For that purpose we construct all possible point pairs $W = \{\{e_i^*, e_j^*\} \mid i, j = 1..|E^*|, i \neq j\}$. Pairs which don't satisfy the angle constrains are discarded. Then the pair with maximum sum of inliers lengths is chosen as the resulting pair of vanishing points $e_h$ and $e_v$.

### 3.5 Algorithm for detection of a vanishing point pair

The Algorithm 1 for vanishing points detection can be summarized as follows:

*Input - the set $S$ of segments on an image.*

1. *Construct the set of rough candidates $E = \{p_{ij} = s_i \cap s_j \mid s_i, s_j \in S, length(s_i) > T_L, length(s_j) > T_L\}$*
2. *For each $e \in E$ evaluate an inliers set $Inl_e = \{s_i \in S \mid D(e, s_i) \leq T_D\}$*
3. *Sort $E$ by a number of inliers*
4. *Discard repeated candidates $e_i \in E$ if $\exists j, 0 \leq j < i : DS_{ij} = Inl_{e_i} \cup Inl_{e_j} \setminus Inl_{e_i} \cap Inl_{e_j}, \sum_{s \in DS_{ij}} length(s) < T_s$*
5. *Construct the set of accurate candidates $E^*$: for each $e \in E$ estimate $e^* = \arg\min_{v \in \mathbb{R}^2} \sum_{s_j \in Inl(e)} \min(T_D, \lambda_{\min}(K_{s_j}(\mathbf{v})))$*
6. *For each $e^* \in E^*$ evaluate an inliers set $Inl_{e^*} = \{s_i \in S \mid D(e^*, s_i) \leq T_D\}$*
7. *Discard repeated candidates from $E^*$*
8. *Discard candidates $e^* \in E^* : dist(e^*, p) > T_d$*
9. *Construct the set of candidate pairs $W = \{w_{ij} = \{e_i^*, e_j^*\} \mid i, j = 1..|E^*|, i \neq j\}$*
10. *Discard pairs $w_{ij} \in W : \alpha_{ij} = \angle \overrightarrow{pe_i^*}, \overrightarrow{pe_j^*}, \alpha_{ij} \leq 90° - T_\alpha^1 \;\|\; \alpha_{ij} \geq T_\alpha^2$*
11. *Return the pair of candidates $w = \{e_h, e_v\} \in W$ with maximum sum of inliers lengths*

In this paper we detect only one pair of vanishing points. However, the algorithm can be easily extended to obtain several alternative pairs. It may be useful if we seek for the rectifications for several objects on a single image (two book pages) or if an image background also contains vanishing points with many segments.

## 4. ESTIMATION OF RECTIFICATION HOMOGRAPHY

Since we have the pair of the orthogonal vanishing points $e_h, e_v \in \mathbb{R}^2$ we can evaluate a homography $H$ that rectifies the source image. For that purpose we construct a new rectifying camera coordinate system $CX'Y'Z'$. Let us denote representations of vanishing points in the homogeneous coordinate system $Oxy$ by $v_h, v_v$. As shown in Section 3.4 the corresponding vectors $V_h, V_v$ are the 3D directional vectors of source horizontal and vertical lines on the object. Then we rotate the camera such as these vectors become collinear to the camera axes. If we determine the new camera x- and y-axes as codirectional to the vectors $V_h, V_v$ the corresponding optical axis may have a reversed direction. It leads to rectified but flipped image. In this case we just invert the axes $CY'$ and $CZ'$.

To calculate the vectors $V_h, V_v$ we should know the camera focal length. If it is not predefined we can calculate it as shown in Section 3.4. However, this problem is ill-conditioned, which results in high inaccuracy of the rectification process. Then we can use an empirical approximation of the focal length: diagonal length of the source image. In this case the vectors $V_h, V_v$ may be not orthogonal. So we calculate a new directional vector of y-axis orthogonal to the obtained x- and z-axes. Then the new camera system is Cartesian but leads to an affine skew with known angle. We can rectify it with an affine transform $A$. Thus the sought metric rectification homography $H$ that transforms the image plane $Oxy$ to a new image plane of the rotated camera can be calculated as $H = KARK^{-1}$ with $R$ being the rotation matrix from old camera system to the new one. The algorithm for estimation of the rectification homography can be summarized as follow.

Algorithm 2

*Input – the pair of vanishing points* $v_h, v_v \in Oxy$

1. $V_h = K^{-1}v_h$, $CX' = V_h / |V_h|$; $\quad V_v = K^{-1}v_v$, $CY' = V_v / |V_v|$; $\quad CZ' = CX' \times CY'$

2. $if\ (CZ'_z < 0):\quad CY' = -CY',\ CZ' = -CZ'$

3. $if\ (CX' \not\perp CY'):\ CY'' = CZ' \times CX';\ else:\ CY'' = CY'$

4. $R = (CX'\quad CY''\quad CZ');\quad \beta = \angle CX', CY';\quad A = \begin{pmatrix} 1 & -\cos(\beta)/\sin(\beta) & 0 \\ 0 & 1/\sin(\beta) & 0 \\ 0 & 0 & 1 \end{pmatrix}$

5. *Return* $H = KARK^{-1}$

Given the set of segments in the image we can obtain the metric rectification homography following the Algorithm 1 for vanishing point detection and Algorithm 2 for homography estimation. It should be noted that the Algorithm 1 does not determine which of two vanishing points is horizontal or vertical. So rectification is estimated up to rotation of a refined image on 90°. In the next section we evaluate the speed and accuracy of the proposed method on real images.

## 5. EXPERIMENTS

While there are many different papers proposing algorithms for rectification images of Manhattan structured objects only a few of them use open datasets to evaluate their methods. For comparison we chose the recent paper [12] that uses neural network (NN) approach and shows the high accuracy of rectification on the open MIDV-500 dataset [20].

To obtain the initial set of segments in the image any method can be used, for example LSD [21]. However, in the experiments we applied the method [22] where 3 types of segments are detected. There were edges which could describe the document borders, ridges which correspond to fill lines, table borders, etc, and text baselines.

### 5.1 Dataset

The MIDV-500 dataset contains 15000 projectively distorted images of planar documents (ID cards) captured using a mobile camera. Many document templates have no inner vertical segments while our algorithm implies the presence of many vertical segments on the object for vertical vanishing point to be found. So we used only following document types: 2, 3, 12 - 17, 19, 23, 26, 30, 31, 36, 38, 39, 49. Also we removed images in which at least one document corners was out of the image borders. It was done to ensure that vertical sides are visible.

The dataset images typically have complex backgrounds such as keyboards, another document sheets etc that may contain many converging segments sets. In our experiments we wanted to evaluate a maximum relative background area (RBA) with which our algorithm still works similar to the NN method. RBA of the dataset images varies from 70% to 90% so we augmented the data to achieve lesser values: 30-60%. For that purpose we cropped images corresponding to the given value γ of RBA. We calculated a cropping rectangle as follows. We obtained a rectangle $R$ with sides parallel to the image borders circumscribed around the document quadrangle. This rectangle was being extended equally to the right and left until the value γ is reached. If one side of $R$ ran into the image border only the other side proceeded to be extended. If the width of $R$ became equal to the image width but a current value of RBA was still less than γ, we similarly extended the rectangle to the up and down.

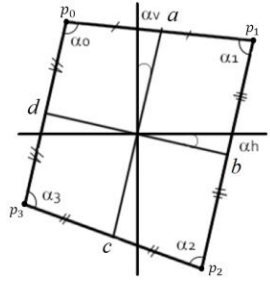

Figure 2. Angles and side lengths explanation for inaccuracy measures $d_{rect}$, $d_{rot}$ and $d_{ar}$

## 5.2 Measure of rectification inaccuracy

Given a document quadrangle $Q$ in a source image, an aspect ratio t of its template and an estimated homography $H$ we calculate three distances for a rectified quadrangle $Q' = HQ$:

$$d_{rect} = \frac{1}{4}\sum_{i=0}^{3}|90°-\alpha_i|, \quad d_{rot} = \frac{1}{2}(|\alpha_v|+|\alpha_h|), \quad d_{ar} = \frac{|(a+c)/(b+d)-t|}{t}$$

The angles $\alpha_i, i=0..3$, $\alpha_v$, $\alpha_h$ and side lengths $a,b,c,d$ of the quadrangle $Q'$ are defined as shown in Fig 2. The distance $d_{rect}$ evaluates the mean deviation of $Q'$ corner angles from 90° and shows how rectangular the document quadrangle is. The distance $d_{rot}$ presents closeness of Q' orientation to the image axes. The $d_{ar}$ distance estimates the deviation in the document aspect ratio. As in the dataset quadrangle corners $p_0, p_1, p_2, p_3$ are given clockwise starting from left top corner we know which sides are horizontal even if image is rotated on 90°. This value can show how much the $Q'$ width differs from the template one if their heights are equal.

## 5.3 Experimental results

With the experiments we wanted to estimate the maximum RBA that allowed our algorithm to reach good rectification results. For that purpose the mean distances were calculated separately for images with RBA from 30% to 60%. For the NN method only angle distances were measured. The authors proposed the algorithm for vanishing points detection and used the simplest method of rectification that did not preserve the aspect ratios. Thus the values of the $d_{ar}$ distance could be random for such rectified images.

Table 1 presents the obtained results of inaccuracy measurements. It can be seen that for the NN method the higher RBA the lower the angle distances values. Probably the reason is that the train dataset contained only real images from MIDV-500 dataset without any augmentation for other RBA values. Because of that we also should take into account the distances values presented in the paper. For the test part of the dataset $d_{rect}=1.71$, $d_{rot}=1.13$.

For images with RBA < 50% the rectification inaccuracy of our method is lower than one of the NN method. With the RBA=50% the results could be seen as roughly the same while higher RBA leads our method to be more inaccurate. Thus we assume that 50% is the maximum relative area of complex background that allowed our method to perform well.

Table1. Measurements of rectification inaccuracy

| RBA | Inaccuracy of our method | | | Inaccuracy of NN method | |
|---|---|---|---|---|---|
| | $d_{rect}$ | $d_{rot}$ | $d_{ar}$ | $d_{rect}$ | $d_{rot}$ |
| 30% | 0.86 | 0.63 | 4.09% | 2.28 | 1.13 |
| 40% | 0.85 | 0.92 | 3.83% | 2.06 | 1.05 |
| 50% | 1.01 | 1.25 | 4.25% | 1.96 | 1.04 |
| 60% | 1.46 | 1.82 | 5.34% | 1.88 | 1.02 |

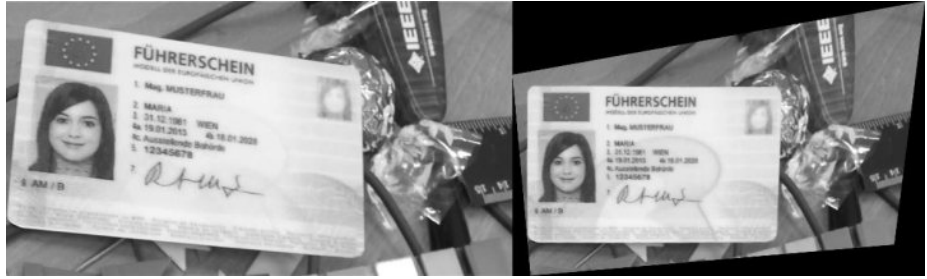

Figure 3. Left: original document image (RBA = 50%), right: the result of document image rectification

From Table 1 it can also be seen that our method stretched rectified document quadrangles by only 4-5%. Such distortion should not result in text rectification errors.

For images with RBA < 50% the rectification inaccuracy of our method is lower than one of the NN method. With the RBA=50% the results could be seen as roughly the same while higher RBA leads our method to be more inaccurate. Thus we assume that 50% is the maximum relative area of complex background that allowed our method to perform well. From Table 1 it can also be seen that our method stretched rectified document quadrangles by only 4-5%. Such distortion should not result in text rectification errors.

Also we measured the mean runtime for both algorithms on core i7 3610qm CPU. For that purpose the proposed method was implemented in C++. It should be noted that the NN method input is an image itself whereas our method employs segments. Therefore we also calculated the mean runtime of the applied segment detection algorithm to compare the time of the whole rectification processes. The runtime of the NN method equals 1.48s, while the time of our method is only 2.23 ms and the time of segment detection is 20.5ms. So the overall process of image rectification is 65 times faster. The example of document rectification with our method is shown in Fig 3.

## 6. CONCLUSION

In this paper we introduced the new method for rectification of projectively distorted Manhattan objects images with complex background which occupies no more than half of the image area. For that purpose we derived the new optimization function of segments for vanishing point position to be estimated. Based on the obtained function the robust method that allows detecting two orthogonal vanishing points was presented. Also we proposed the original method that estimates camera rotation and corresponding rectification homography with vanishing points.

According to the experiments the proposed method shows rectification quality as good as the state-of-the-art method on the distorted document images dataset MIDV-500. The inaccuracy in length ratio preservation is around 5%. Experimental results also show that the method runtime is near 3ms on core i7 3610qm.

## ACKNOWLEDGMENTS

This work is partially supported by Russian Foundation for Basic Research (projects 18-29-26035 and 17-29-03236).